\documentclass{article}

\usepackage{arxiv}

\usepackage[utf8]{inputenc} 
\usepackage[T1]{fontenc}    
\usepackage{hyperref}       
\usepackage{url}            
\usepackage{booktabs}       
\usepackage{amsfonts}       
\usepackage{nicefrac}       
\usepackage{microtype}      
\usepackage{lipsum}  
\usepackage{authblk}
\usepackage{graphicx}
\usepackage{natbib}
\usepackage{doi}
\usepackage{multirow}
\usepackage{subfigure}
\usepackage{caption}
\usepackage{amsmath}
\usepackage{cleveref}       

\title{Large Language Models as Master Key: Unlocking the Secrets of Materials Science with GPT}

\date{April 5, 2023}

\author[1,2,*,**]{Tong Xie}
\author[2,3*]{Yuwei Wan}
\author[6]{Wei Huang}
\author[3]{Yufei Zhou}
\author[2,6] {Yixuan Liu}
\author[2,6]{Qingyuan Linghu}
\author[1,2]{Shaozhou Wang}
\author[3]{Chunyu Kit}
\author[4,5]{Clara Grazian}
\author[6]{Wenjie Zhang}
\author[1,**]{Bram Hoex}

\affil[1]{School of Photovoltaic and Renewable Energy Engineering, University of New South Wales, Kensington, NSW, Australia}
\affil[2]{GreenDynamics Pty. Ltd, Kensington, NSW, Australia}
\affil[3]{Department of Linguistics and Translation, City University of Hong Kong, Hong Kong, China}
\affil[4]{School of Mathematics and Statistics, University of Sydney, Camperdown, NSW, Australia}
\affil[5]{DARE ARC Training Centre in Data Analytics for Resources and Environments, Australia}
\affil[6]{School of Computer Science and Engineering, University of New South Wales, Kensington, NSW, Australia}
\affil[7]{School of Computing and Information Technology, University of Melbourne, Parkville, VIC, Australia}
\affil[*]{Authors contributed equally}
\affil[**]{Corresponding author: tong.xie@unsw.edu.au, b.hoex@unsw.edu.au}

\renewcommand{\headeright}{}

\hypersetup{
pdftitle={A template for the arxiv style},
pdfsubject={q-bio.NC, q-bio.QM},
pdfauthor={David S.~Hippocampus, Elias D.~Striatum},
pdfkeywords={First keyword, Second keyword, More},
}

\begin{document}
\maketitle

\begin{abstract}
The amount of data has growing significance in exploring cutting-edge materials and a number of datasets have been generated either by hand or automated approaches. However, the materials science field struggles to effectively utilize the abundance of data, especially in applied disciplines where materials are evaluated based on device performance rather than their properties. This article presents a new natural language processing (NLP) task called structured information inference (SII) to address the complexities of information extraction at the device level in material science. We accomplished this task by tuning GPT-3 on an existing perovskite solar cell FAIR (Findable, Accessible, Interoperable, Reusable) dataset with 91.8\% F1-score and extended the dataset with data published since its release. The produced data is formatted and normalized, enabling its direct utilization as input in subsequent data analysis. This feature empowers material scientists to develop models by selecting high-quality review articles within their domain. Additionally, we designed experiments to predict the electrical performance of solar cells and design materials or devices with targeted parameters using large language models (LLMs). Our results demonstrate comparable performance to traditional machine learning methods without feature selection, highlighting the potential of LLMs to acquire scientific knowledge and design new materials akin to materials scientists.
\end{abstract}


\section{Introduction}
DData has since long been the cornerstone of empirical science, serving as the foundation upon which discoveries are made, and our understanding of the world is built. In recent years, big data has become an indispensable resource for various industries, particularly the technology sector. Materials science is no exception to this trend. It has revolutionized the exploration and design of cutting-edge materials for a wide array of applications, encompassing catalysts \cite{Jensen2019}, thermoelectrics \cite{Tshitoyan2019}, and batteries \cite{Huang2022a,Huang2022} solutions. These initiatives underscore the growing significance of data in materials research, paving the way for groundbreaking innovations in the field.

Nonetheless, despite the broad acknowledgement of data’s significance and the ongoing initiatives to exploit its potential, the experimental materials science domain persistently encounters difficulties in effectively utilizing the abundance of available data \cite{2017EmptyScience}. This problem is especially evident in applied disciplines, where materials are frequently evaluated predominantly based on their device performance rather than a thorough comprehension of their inherent properties and behaviour \cite{Olivetti2020}. A crucial issue in this context is proficiently extracting relevant information from the extensive, unstructured scientific literature. This challenge not only impedes the comprehensive understanding of material candidates and their attributes but also impedes the recognition of future applications. It also contributes to the bottleneck in the materials discovery pipeline, wherein experimental synthesis remains laborious and time-consuming.

Natural language processing (NLP) techniques, especially named entity recognition (NER), have proved promising in addressing these issues, offering the potential to streamline the extraction of relevant information from the scientific literature. However, most importantly, they still cannot perform information extraction at the device level since one device has complex relations between each material and entity. In this study, we introduce a new NLP task called structured information inference (SII) to leverage pre-existing review paper datasets in materials science. This task is at the discourse level and, in practice, covers mainstream tasks such as NER, entity resolution (ER), relation extraction (RE), and information inference (II). We accomplished this complex task in one step by fine-tuning GPT-3 on a manually summarized perovskite solar cell FAIR (Findable, Accessible, Interoperable, Reusable) dataset published in February 2021 \cite{Jacobsson2022}. Our fine-tuned model not only exhibits high precision but also possesses the capacity to update all scientific datasets derived from review papers automatically. We demonstrated this novel approach on the very active field of perovskite solar cells and successfully extracted intricate relationships, and constructed a graph network of other perovskite solar cells device-level knowledge from March 2021 to March 2023. Our approach provides evidence that LLM can autonomously learn complex knowledge data frames and construct output as predefined schema from review papers without additional manual annotation. The produced dataset is formatted and normalized, enabling its direct utilization as input in subsequent data analysis (like machine learning) without processing steps. This feature will enable materials scientists to develop their own models by merely selecting high-quality review papers within their domain. Additionally, the LLM API can be directly employed on personal computers to accomplish a wide range of intricate materials informatics tasks, supplanting the costly post-training process of custom and localized language models.

\section{Related Work}

In material science, significant effort has been devoted to extracting entities such as chemical terminologies, properties and synthesis parameters in related scientific literature. The related NER methods can be roughly classified as rule-based, recurrent neural network (RNN), and transformer-based large language model (LLM). Rule-based methods are usually unsupervised and rely on dictionaries or regex rules (like extracting words formed by chemical elements and numbers) summarized by experts. ChemDataExtractor 1.0 \cite{Swain2016} is a traditional text mining tool, using rule-based methods to extract chemical information from scientific documents and some databases \cite{Huang2020, Beard2019} were auto-generated using this tool. In contrast to rule-based approaches, recurrent neural network (RNN) models trained on annotated data exhibit enhanced flexibility in handling a diverse range of entity types, such as chemical compounds, proteins, genes, materials, and synthesis methods. A representative RNN method, BiLSTM-CRF \cite{Huang2015}, has been applied to extract chemical synthesis parameters in the methods section \cite{Weston2019, Kononova2019, He2023}. These studies linked material names with their co-occurring entities to generate synthesis recipes or analyze potential relations. In recent years, the advent of LLMs such as BERT \cite{Devlin2018} have become the state-of-the-art of many NLP tasks, including NER. Both fine-tuned BERT \cite{Zhao2021} and domain-specific pre-trained BERT \cite{Walker2021, Huang2022a} have shown substantial improvement in material science NER tasks compared to RNN methods. Trewartha et al. \cite{Trewartha2022} made a systematic comparison and concluded that the performance of domain-specific pre-trained BERT has better performance than domain-specific enhanced BiLSTM. Several datasets \cite{Sierepeklis2022AChemDataExtractor, Dong2022Auto-generatedChemDataExtractor, Beard2022Perovskite-ChemDataExtractor} employ NER to produce automatically generated tabular databases of material property data aggregated from textual entries. Despite these endeavours, a significant portion of experimental materials science still faces challenges in effectively utilizing the generated data, particularly in applied fields where materials are predominantly assessed based on their performance in devices. This necessitates the extraction of relationships between textual entries.

Extracting relationships between entities in materials science has always been a challenge and has received relatively less attention than the NER task. Mysore et al. \cite{Mysore2019} built a dataset of 230 synthesis procedures with labelled graphs whose nodes are synthesis operations and their typed arguments, and labelled edges specify relations between the nodes. MatSciBERT \cite{gupta2022} yields the best performance of RE on this dataset. Most existing research treats RE as the step following the NER in an information retrieval pipeline and usually focuses on intra-sentence binary relationship \cite{Pawar2017}. As a result, RE is essentially a classification task for pairs of recognized entities in these pipeline-based systems \cite{Song2015, Cejuela2018}. Nonetheless, real-world situations are considerably more intricate than this elementary context. Recent advancements in RE research are focused on addressing more sophisticated RE scenarios, such as those involving multiple sentences \cite{Pawar2017}. It is worth noting that N-ary relations (N entities are involved in a relation), with more challenges, have received increasing attention. Recently, \cite{Dunn2022} propose a sequence-to-sequence large language model (seq2seq-LLM) approach capable of addressing complex interrelations without the need to enumerate all possible N-ary relations. They fine-tuned GPT-3, a prominent LLM developed by OpenAI \cite{Brown2020}, to jointly perform NER and RE tasks, aiming to extract hierarchical entity relationships. 
 \begin{figure}[h!]
  \centering
  \includegraphics[width=\linewidth]{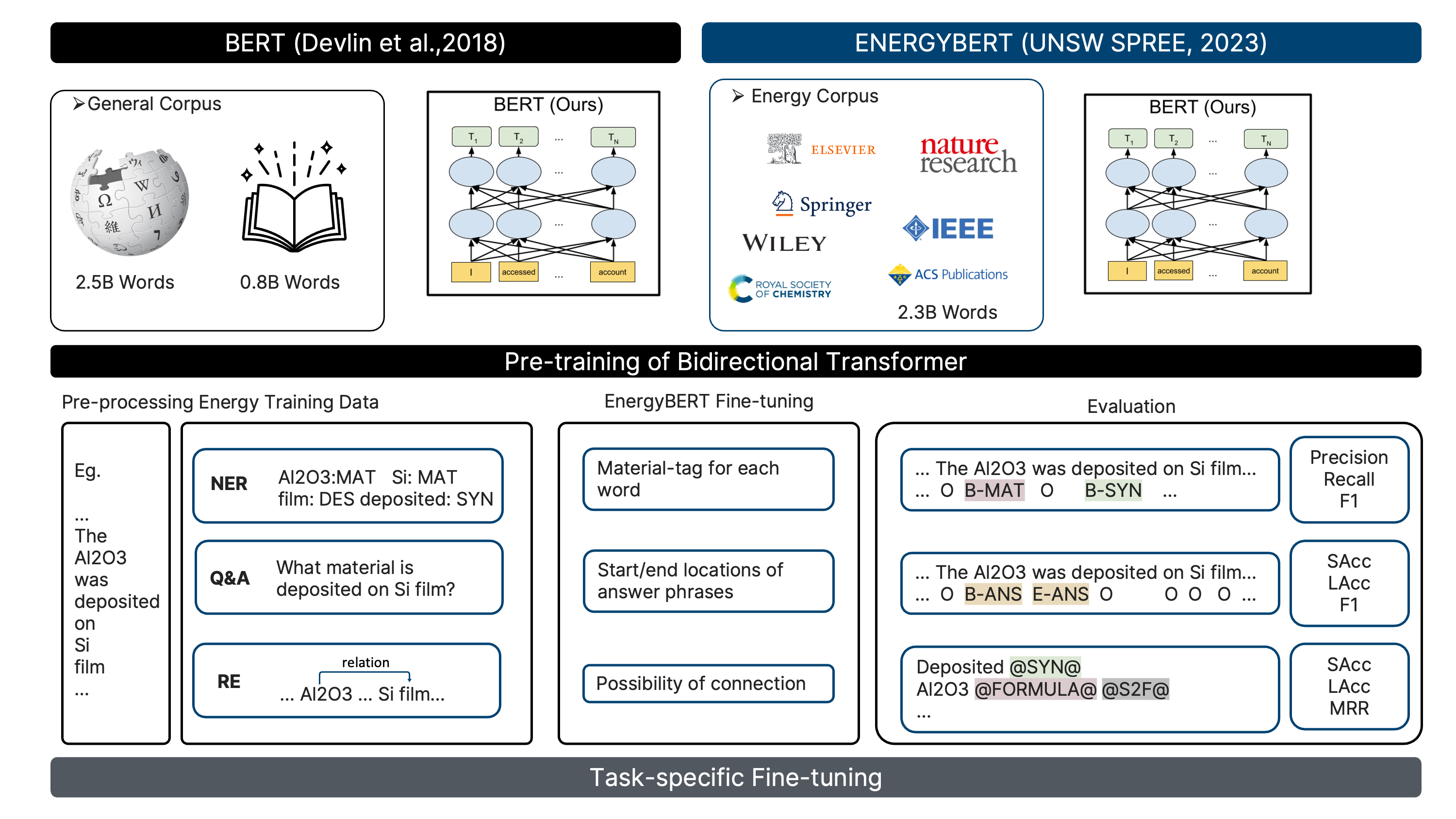}
  \caption{Overview of the pre-training and fine-tuning of BERT \cite{Devlin2018} and EnergyBERT}
\end{figure}

\textbf{Issues of traditional annotation mechanism}

The research mentioned above used a word-by-word traditional annotation mechanism (see Figure 1), which does not align with the needs of material scientists. On the one side, this mechanism typically demands substantial effort from both natural language processing and materials science experts in several aspects: 1) creation of NER categories; 2) development of a labelling interface; 3) learning costs associated with NER/RE labelling rules; and 4) time costs of NER/RE labelling. Conversely, scientific information often transcends simple pairwise relationships between entities. For instance, a compound's properties are determined by multiple factors, such as material name, phase structure, morphology, and synthesis methods. This complexity is exemplified in the distinction between 'PECVD(plasma-enhanced chemical vapour deposition) Al-doped TiO\textsubscript{2} film' and 'ALD (Atomic Layer Deposition) Al-doped TiO\textsubscript{2},' which exhibit different properties. Additionally, materials knowledge is frequently hierarchical, with relations that may only be valid between one entity type and a compound entity comprising multiple entities and relationships. Theoretically, such relations can be modelled as N-ary, but comprehensively enumerating all possible variations is both impractical and unsuitable for traditional relation extraction methods, as each relation type necessitates a sufficient number of training examples.

Due to the difficulty in annotation and simulation of material information, high-quality annotated data is limited throughout material science, prompting us to utilize existing review paper databases. A review article represents a scholarly publication that amalgamates and evaluates prior research papers on a specific topic. Such articles investigate distinct research questions or theoretical or practical approaches, providing readers with an extensive and up-to-date understanding of the research area. These articles contain natural, high-quality summaries and intricate relationships in domain-specific subjects, materials and properties, and device information. We endeavoured to retrace the summarized information in the review paper, as provided by other scientists, back to the original text through entity annotation and relation annotation. However, these efforts did not result in a perfect match with the corresponding sections. According to statistics, exact match demonstrates a low matching rate of 44.7\% on average. The unmatched parts need exports to annotate manually, while the annotation time per material entry is estimated to be 20 seconds \cite{Dunn2022}. Thus, converting this dataset into traditional word-by-word NERRE data annotation proved to be challenging.

\textbf{Opportunities and new NLP task - Structured Information Inference}

As discussed in previous sections, existing research primarily concentrates on identifying entities and their relationships. However, the practical process of extracting information by materials scientists is considerably more complex. The expressions found in research articles are intricate and varied. At the device level, not only do complex relationships between materials need to be considered, but units may also differ across publications. For example, '0.3 cm2' is equivalent to '0.003 mm2'. Entity definitions can be flexible, and occasionally their meanings depend on words in separate paragraphs. A paper might only mention 'Al-doped TiO\textsubscript{2}', leaving scientists to infer whether it is 'ALD' or 'PECVD' grown. Furthermore, field-specific vocabulary may introduce ambiguity, complicating matters further; for instance, 'Al-doped TiO\textsubscript{2} film' could be synonymous with 'Al\textsubscript{x}Ti\textsubscript{y}O film'. We examined various phenomena that scientists employ to summarize and infer information from materials science articles and discovered that these phenomena could be aligned with existing NLP tasks, as demonstrated by the examples in Fig. 2(b):

\begin{itemize}
    \item Named entity recognition (NER): Direct information extraction such as material names and temperature.
    \item Entity Recognition (ER): Standardizing the expression format, units, abbreviations, etc. of the information. 
    \item Information inference (II): Processing and reasoning with the information. 
    \item Relationship Extraction (RE): Discerning the connections between individual entries or groups of entries.
\end{itemize}
Extracting information from scientific texts could be more challenging than the simple process of general texts. Moreover, a piece of material knowledge might be inferred through multiple NLP tasks with multiple entities. For instance, the  Al-doped TiO\textsubscript{2} compact layer could be inferred as ALD c-Al\textsubscript{x}Ti\textsubscript{y}O layer in a review paper or FAIR dataset. (when the deposition method is mentioned in another paragraph). These complexities make annotating related training datasets particularly demanding, especially at the documents level, as they represent an accumulation of materials science knowledge spanning centuries. To simulate the process of scientists extracting information from domain-specific texts, we propose a new NLP task designed for the scientific field called structured information inference (SII). This task aims to jointly information inference(or extraction) and relationship extraction. The relationship could be hierarchically or as a list of multiple items without enumerating all possible n-tuple relationships. Initially, we attempted to use BERT-based approaches. Still, the need to determine specific tasks for each piece of information complicated the problem, rendering the original BERT or domain-specific BERT unsuitable for fine-tuning. However, the advent of GPT-3 and its related applications offers new opportunities. As depicted in Figure 2(a), GPT-3 and other generative language models employ a decoder structure, which is well-suited for seq-to-seq tasks (i.e., input text generates output text) and aligns with the generating logic of material scientists in collecting data from the literature. Consequently, we propose fine-tuning GPT-3 \cite{Brown2020} to infer key information from original papers directly. This approach not only saves significant time and cost but also leads to more accurate and comprehensive information summarization. The GPT-3 model can capture high-dimensional information and relationships within a paper that traditional word-by-word labelling mechanisms may overlook.

\begin{figure}
  \centering
  \includegraphics[width=\linewidth]{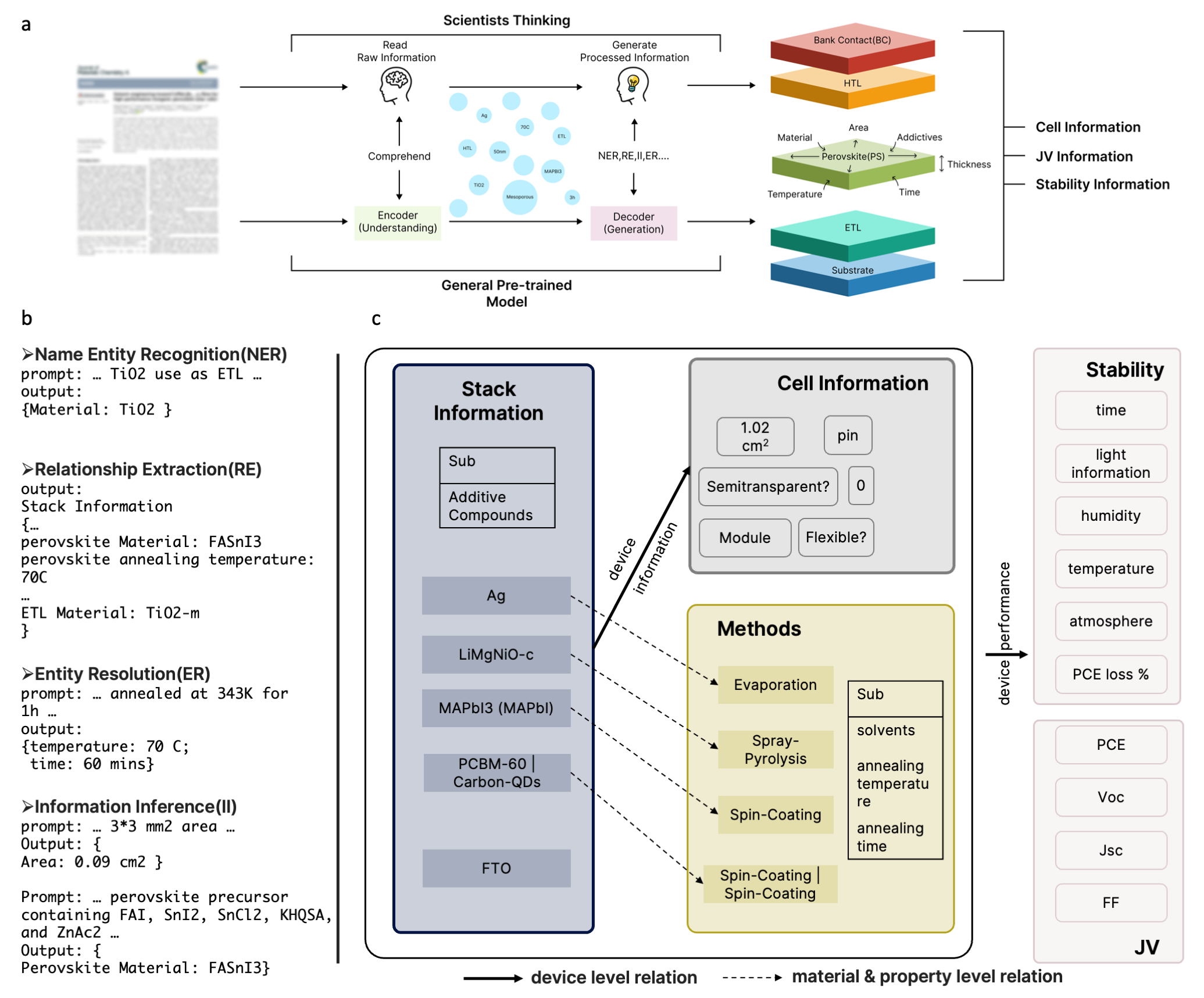}
  \caption{(a) An overview of structured information inference through multi-task learning is presented, with the decoder responsible for comprehending tasks and generating the corresponding outputs. (b) Examples of generation tasks include named entity recognition, information inference, entity resolution, and relationship extraction. (c) A complex, multi-dimensional graph network of perovskite solar cells utilized in this study is constructed based on the framework proposed by Jacobsson et al.  \cite{Jacobsson2022}}
\end{figure}

\begin{figure}
  \centering
  \includegraphics[width=\linewidth]{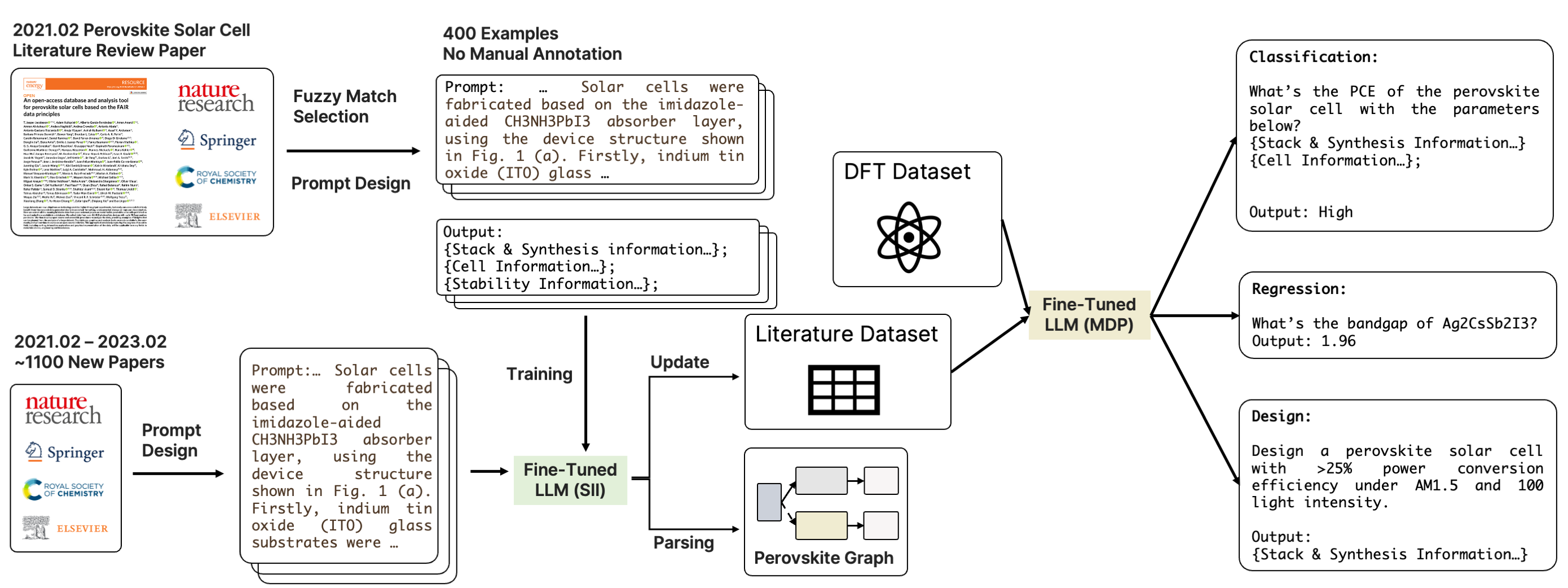}
  \caption{Flowchart of the fine-tuned LLM for structured information inference(SII) tasks and material \& device prediction(MDP) tasks}
\end{figure}

\section{Methods}

\textbf{Dataset background and Graph Network Design}

This study's database contains over 1.2 million full-text research articles on material science and energy topics, sourced from reputable publishers such as Elsevier, Springer Nature, and the Royal Society of Chemistry. These articles were provided in XML/HTML format. From the 42,000 device data records manually extracted by Jacobsson in 2022, we associated them with more than 15,000 corresponding articles, focusing exclusively on English-language publications. The original dataset encompassed 95 attributes, covering stack information, system-level data, and performance metrics. This information was then integrated into a graphical network, exemplified in Figure 2(c).
The information was organized into two levels: material \& property level information and device level information (details can be found in Appendix B).

\textit{Material \& Property Level Information:} a. stack information (Set \emph{A}) and methods information (Set \emph{B}) for each layer, encompassing substrate, electron transport layer (ETL), perovskite, hole transport layer(HTL), and back contact.

\textit{Device Level Information:} a. stability (Set \emph{C}) and electrical (J-V) performance data (Set \emph{D}).

While the original dataset possesses the capacity to capture more than 400 attributes, our study focuses on those with abundant information and widespread usage. We further categorized the relationships between each entity into two levels: material and property or device connections.

\textbf{Schema Design}

In fine-tuning, we transformed the original tabular data into a plain text schema to facilitate the model's understanding. Each schema is presented as a dictionary, where the keys represent the element names, and the values represent the corresponding elements. For each relevant paragraph, we aimed to enable the model to learn how to automatically and accurately summarize a corresponding schema. To compare with the direct use of GPT-3.5, we conducted prompt design to obtain results that are as close to the original format as possible (We opted not to utilize the original GPT-3 for comparison, as even with well-designed prompts, the model produced results that were exceedingly difficult to decipher, often containing numerous repetitions and nonsensical outputs.). After multiple attempts, we designed a prompt using the original paragraph with the prefix ``Read the following paragraphs and extract the information below:'' and list of element names attached. We removed underscores in element names and added some requirements to limit the length or content of the element names. For example, we added ``(only name, not details)'' after the element name ``HTL deposition procedure''. For element names that require boolean answers, we changed the element names into general questions. For example, we changed ``Module'' to ``Any Module test?''.  

\textbf{Dataset preparation (fuzzy match mechanism)}

For the dataset preparation, we first process the full text of scientific papers and then propose a fuzzy match mechanism to calculate the match rate between the schemas and the underlying text. To process the scientific papers, we split each full text into different sections with headers and content. To meet the 2049 token limit requirement (about 1500 words) of GPT-3, we only extracted sections that contained keywords `experimental', `materials', `methods' or `experiment' in the section's header as input, which was found to be the most informative sections. 

After extracting the most informative sections of each scientific paper, we applied the fuzzy match mechanism to calculate the match rate between each pair of the schemas and the extracted content within the sample set. Then the samples were ranked according to the calculated match rate. For the fuzzy match mechanism, the match rate of a given schema with the underyling content was calculated based on the number of each key-value pair within the schema that matches the underlying extracted content over the total number of key-value pairs. For more detail, the value of each key in a given schema is first split into a list of values by delimiters (e.g., ‘|’, ‘;’,‘:’), and ‘spin-coating’ in the schema was replaced by ‘spin-coated’. After that, each key will have a given matching rule. For a given key, if one of the values in the value list matches the underlying content according to the given matching rule, then we consider this key-value pair matched the underlying content. For the key `ETL\underline{~}stack\underline{~}sequence', if the whole string or the substring before `-' in a given value appears in the original content, then the value matches the content. For the keys `Perovskite\underline{~}composition\underline{~}long\underline{~}form' and `Perovskite\underline{~}composition\underline{~}short\underline{~}form', if a given value is a subset of a single word within the original content, then the value matches the content. For the keys with the value `unknown', we always count them as a match. For all other keys, if the value appears in the original content, then the value matches the original content. For learning efficiency, we rank the schemas of each underlying content by the match rate and only select the top 1 schema of each underlying content as the sample. We then rank the samples by schema match rate and select the top 400 samples as the training sample. The match rates of these selected samples ranged from 100\% to about 85\%.

\textbf{Fine-tuning details}

We choose davinci (175B parameters) \cite{Brown2020} as our base model since it is the most capable GPT-3 model available for fine-tuning. Each data sample has a prompt and a completion. Specifically, the prompt is the text extracted from scientific papers with several paragraphs having schema information. And the completion act is the answer to those schemas, including 31 key-value pairs in the form of ``schema name: answer''. As recommended by the OpenAI CLI tool, our dataset is transferred into .jsonl format, where ``\textbackslash n'' is inserted at the end of the prompt and `` '' (space) at the beginning of the completion. Then the dataset is split into a training set and a test set containing 360 and 40 samples separately. The model is trained for 4 epochs at a batch size of 1 with a 0.1 learning rate multiplier and 0.01 prompt loss weight the via OpenAI API.

\textbf{Evaluation}

We evaluated the performance of the fine-tuned model and GPT-3.5 on SII task using four decomposed sub-tasks, NER, RE, ER and II. The metrics of the NER task evaluated how likely an output schema (prediction) was matched with the target schema (answer). Each element in the output schema can be seen as an entity $E^{p}$ and the corresponding element in the target schema can be seen as an entity $E^{a}$. We design a word-basis measurement by separating an entity $E$ into a set of words $S = \{w_1 , w_2 , w_3 , ..., w_k \}$ and comparing the difference between $S^{p}$ and $S^{a}$. The separators include ``;", ``|", ``:" and ``>>". After separating both entities, the number of matching words in both sets is counted as true positives ($S^{p} \cap S^{a}$) and the set difference is counted as false positives 
($S^{p} \setminus S^{a}$) or false negatives ($S^{a} \setminus S^{p}$). For example, an element in the output schema is ``70.0 >> 120.0'' and the corresponding answer was ``70.0 >> Unknown'', and we recorded one true positive ``70.0'', one false positive ``120.0'' and one false negative ``Unknown''. With true positives ($tp$), false positives ($fp$) and false negatives ($fn$) identified, metrics of each pair of entities were calculated as:

\begin{align}
\mbox{precision} &= \frac{tp}{tp+fp} \\
\mbox{recall} &= \frac{tp}{tp+fn} \\
{\mbox{F1-score}} &= \frac{2\times \mbox{precision} \times \mbox{recall}}{\mbox{precision}+\mbox{recall}} 
\end{align}

The metrics of the RE task evaluate how likely the output schema caught the inner relationship between related element sets. According to the nature and internal relation of elements in the schema described in Section 3.1, we construct three types of relations: \emph{A-B}, \emph{A-C}, and \emph{ABC-D}. The relationships are also scored by a word-basis measurement similar to the one NER uses, using a number of correct collocations. Each collocation is a n-tuples relating words $w^m_n$ of each involved entities $E_n$ in relation $r$. For each type of relation, we can summarize collocations in the predicted schema into a predicted relation set ($R^{p}$), and those in the answer schema into an answer relation set  ($R^{a}$). The number of matching collocations in both relation sets is counted as true positives ($R^{p} \cap R^{a}$) and the collocation difference is counted as false positives ($R^{p} \setminus R^{a}$) or false negatives ($R^{a} \setminus R^{p}$). After all kinds of collocations were identified, metrics of relation extraction were calculated with the same Equations (1), (2), and (3) described above.

In addition to word-basis measurement, we also manually evaluated the performance of NER and RE. Two experts with domain knowledge of material science were invited to manually judge the prediction of models by their quality. For each prediction of the element, they need to give a score from 0 (incorrect), 1 (correct but with unrelated information), and 2 (correct). When they have different opinions on the same prediction, they negotiate with each other and give a final decision. Both 1 and 2 were counted as correct to calculate the accuracy of manual evaluation. However, there is a discrepancy in the evaluation scores: when using the exact match to evaluate, it is too strict for GPT-3.5 without format training, while manual evaluation ignores the form differences and may not be fair to the fine-tuned model (not counting its ability of ER and II). Thus, we further evaluate the performance of II, ER-U (Entity Resolution for Units), and ER-T (Entity Resolution for Terminology). The elements selected do not appear in the original text, which means their target answers have changes in form, scale, or expression compared to corresponding parts in the original text (see examples in Table 3). Only an exact match with the target answer is counted as correct.

\section{Results}

We will report the results in four parts: NER results, RE results, ER and II results, as well as the effect of the training set size. The results show that our fine-tuned model outperformed GPT-3.5 in both NER and RE tasks, and also obtained high accuracy in unique ER and II tasks. Finally, we examine the effect of the training set size on training token accuracy. This section demonstrates the necessity of domain-specific fine-tuning and provides readers with a more comprehensive understanding of the model's performance. 

\textbf{NER results}

Table 1 shows the results for the schema elements matching computed by metrics described in Section 3.5 as well as human evaluation. It is known that GPT-3.5 is more powerful than GPT-3, but our model fine-tuned on GPT-3 surprisingly outperformed GPT-3.5 by 62.9 points in F1-score and 22 points in the manual evaluation. As for NER results of different sets, it can be observed that elements in the set \emph{D} are more challenging for both models (59.3\% and 89.3\% in manual evaluation).

Without fine-tuning, the elements generated by GPT-3.5 are longer than the target answers, especially in procedure-related elements (even though we have attempted to impose length restrictions during prompt design). According to statistics in manual evaluation, about 15\% of correct predictions produced by GPT-3.5 have significant unrelated information, while this ratio is only 4\% for our fine-tuned model. Consequently, GPT-3.5 had a lower precision and higher recall considering the averaged length of the output. However, the recall of GPT-3.5 is still significantly lower than a GPT-3 fine-tuned on material scientific knowledge datasets, which implies that GPT-3.5 cannot accurately summarize hidden information in input paragraphs directly. In contrast, a fine-tuned model not only finds the corresponding parts accurately, but also learns to summarize, normalize or even deduct.

\begin{table}[h!]
\centering
    \caption{Results of NER in SII through multi-task learning}
    \begin{tabular}{cccccc}
    \toprule
    Model & Set & Precision & Recall & F1-score & Manual\\
    \midrule
    \multirow{5}*{GPT-3.5} & \emph{A} & 19.9 & 44.9 & 27.6 & 71.0\\
		  & \emph{B} & 13.1 & 27.1 & 17.7 & 72.1\\
            & \emph{C} & 18.9 & 57.3 & 28.4 & 82.5\\
            & \emph{D} & 27.0 & 43.6 & 33.3 & 59.3\\
    \cline{2-6}
            & total & 22.6 & 43.0 & 28.7 & 72.1\\
    \midrule
    \multirow{5}*{Fine-tuned model} & \emph{A} & 93.1 & 91.9 & 92.5 & 96.0\\
		  & \emph{B} & 96.1 & 96.1 & 96.1 & 96.3\\
            & \emph{C} & 89.8 & 87.4 & 89.0 & 93.6\\
            & \emph{D} & 89.3 & 89.3 & 89.3 & 89.3\\
    \cline{2-6}
            & total & 92.4 & 91.4 & 91.8 & 94.1\\
  \bottomrule
\end{tabular}
\end{table}

\textbf{RE results}

Table 2 reports the RE scores, which evaluate the consistency of inner element sets of output schema. Since a proper relation must be based on the correct extracted entities, the performance of the RE task is influenced by the performance of the NER task. Thus, the RE scores here can be seen as a reflection of the NERRE task, not RE merely. 

Overall, the fine-tuned model significantly outperforms GPT-3.5 in all three types of relation extraction. Specifically, the performance degradation of the fine-tuned model (about 5\%) from the NER to NERRE task is much smaller than that of GPT-3.5 (about 20\%). The difference between precision and recall is also smaller (about 4\%) for the fine-tuned model. Among the three types of relations, the performance of relation \emph{ABC-D} is relatively the worst (lowest F1-score), which may be attributed to poorer NER performance of the d set and more involved entities. In the \emph{ABC-D} relation extraction, the fine-tuned model has notably obtained a comparatively lower F1-score, but a comparable score to other types in manual evaluation. In comparison, the fine-tuned model has a more balanced performance among the three types of relations.

\begin{table}[h!]
\centering
    \caption{Results of RE in SII through multi-task learning}
    \begin{tabular}{cccccc}
    \toprule
    Model & Relation & Precision & Recall & F1-score & Manual\\
    \midrule
    \multirow{3}*{GPT-3.5} & \emph{A-B} & 5.02 & 11.96 & 6.67 & 43.4\\
		  & \emph{A-C} & 7.23 & 29.51 & 10.3 & 66.5\\
            & \emph{ABC-D} & 2.76 & 10.73 & 3.95 & 49.38\\
    \midrule
    \multirow{3}*{Fine-tuned model} & \emph{A-B} & 90.54 & 88.46 & 89.39 & 88.0\\
		  & \emph{A-C} & 84.73 & 80.87 & 82.33 & 90.8\\
            & \emph{ABC-D} & 71.39 & 67.12 & 68.49 & 87.6\\
  \bottomrule
\end{tabular}
\end{table}

\textbf{II and ER results}

Table 3 shows the support number and accuracy of II, ER-U (entity resolution for units) and ER-T (entity resolution for terminology) on our finetuned model, respectively. To help to understand, we also give the example prompts and outputs of them. For these tasks, we did not display the results of the GPT-3.5 model because the accuracy of each task is 0\% or close to 0\%. In comparison, the high accuracy achieved by fine-tuning the model indicates that it can not only complete the ER and II tasks but also perform well. This suggests that fine-tuning can greatly enhance the ability of the GPT-3 model to comprehend data formats and fill in missing information, simulating the process by which scientists extract and process data from research papers.

\textbf{Effect of the training dataset size}

The relationship between elapsed tokens (the number of tokens the model has seen so far) and training token accuracy (the percentage of tokens in the training batch that the model correctly predicted) is plotted in Figure 4(a). The relationship between elapsed examples (the number of examples the model has seen so far) and training loss (loss on the training batch) is plotted in Figure 4(b). Since we set the number of epochs to be equal to 4, the examples above 360 are repeated. It can be observed that there is a sharp increase in training token accuracy and a sharp reduction of training loss during the first 50 examples, which implies that GPT-3 can learn the process of producing schema very fast within only a few schemas shown to it. Above 100 examples, the performance improvement brought by increasing the amount of data is relatively slow and marginal.

\begin{table}[h!]
\renewcommand{\arraystretch}{2}
    \caption{Results of II and ER in SII through multi-task learning }
    \centering
    \begin{tabular}{cccccp{1.1cm<{\centering}}}
    \toprule
    Task & Example Prompt & Example Completion & Support & Accuracy\\
    \midrule
    \centering II & \centering \shortstack{ … perovskite precursor \\ containing FAI, SnI2 … } & Pervoskite Material : FASnI3 & 61 & 91.80\\
    \midrule
    ER-U & \shortstack{ … annealed at 343K for 1h … } & \shortstack{temperature: 70 C; time: 60 mins} & 65 & 69.23\\
    \midrule
    ER-T & \shortstack{ … mesoporous TiO\textsubscript{2} … } & \shortstack{Material: TiO\textsubscript{2}-m} & 117 & 87.18\\
  \bottomrule
\end{tabular}
\end{table}

\begin{figure}[h!]
\centering
\subfigure[Relationship between elapsed tokens and training token accuracy]{\label{fig:a}\includegraphics[width=80mm]{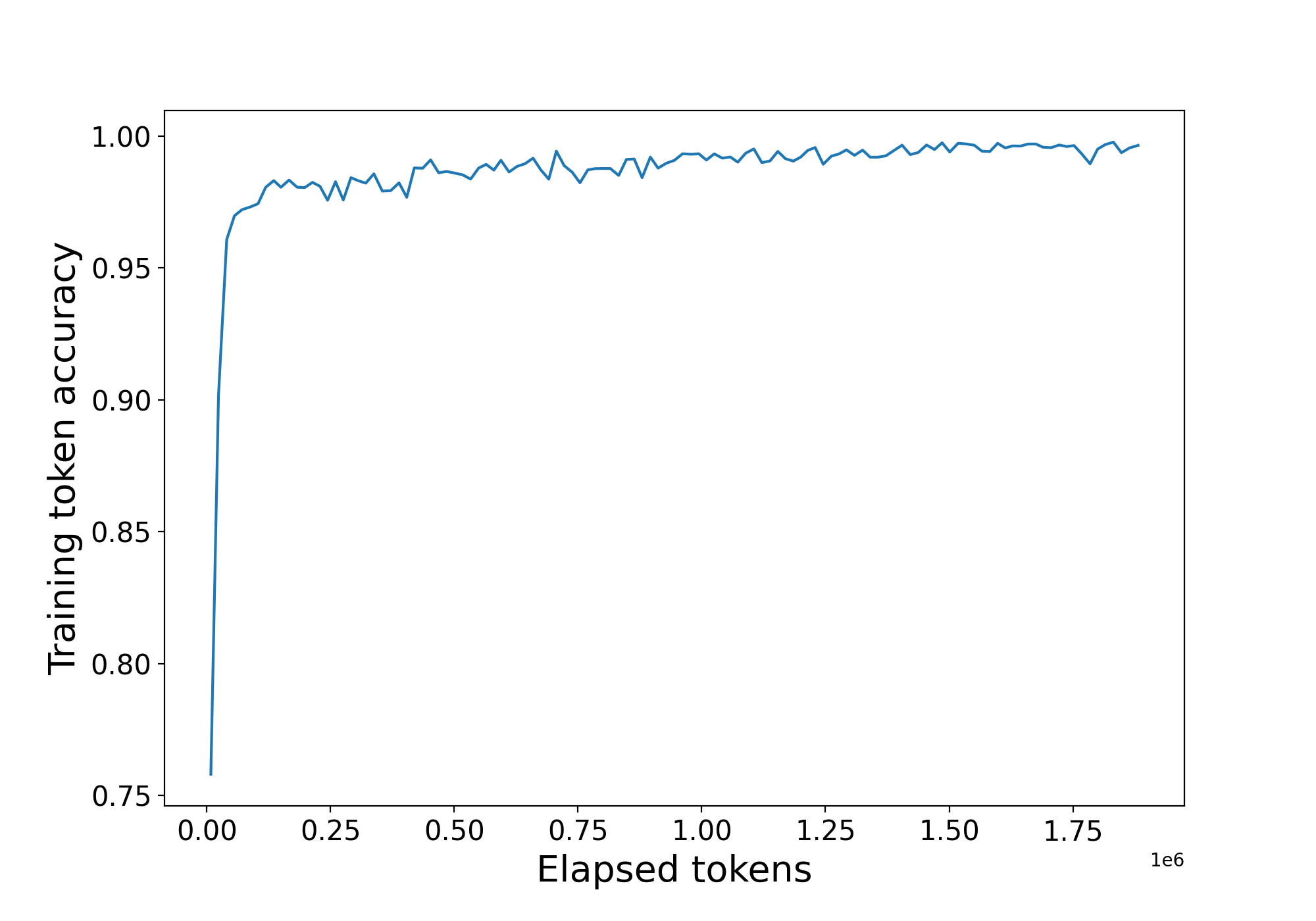}}
\hfill
\subfigure[Relationship between elapsed examples and training loss]{\label{fig:b}\includegraphics[width=80mm]{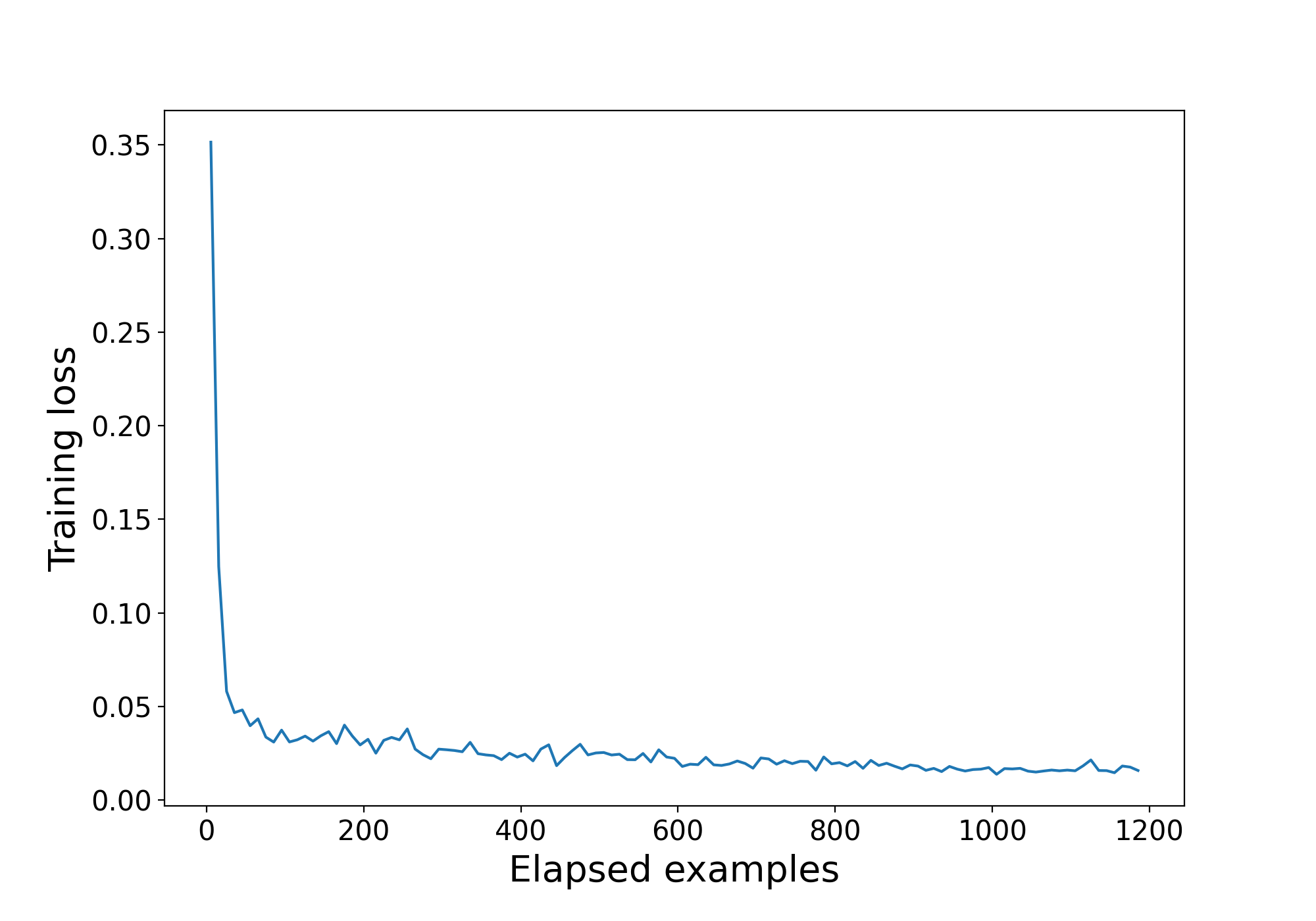}}
\caption{Effect of training data size on SII task for perovskite solar cell}
\end{figure}

\section{Discussion}

\textbf{Comparison of performance}

Based on the predicted schemas, we summarize the issues of direct use of GPT-3.5: 1) Unstable generation: the predicted schema sometimes misses one or two elements (we consider these missing elements as incorrect answers during evaluation). 2) The suggested schema can also change the expression of the generated answer, for example, ``Backcontact additives compounds'' becomes ``Backcontact additives/compounds''. 3) There can be multiple expressions for the same answer, such as ``not mentioned'', ``N/A'', ``None mentioned'', and ``not specified'', which causes difficulties in parsing and predicting a unified format. 4) Variable length: the generated answer's length is not fixed and can sometimes be very long, even if the prompt design limits the length (the limit not always works). 5) Unnecessary changes in expression for correct answers, such as ``60 min'' being changed to ``1 hour''. 6) Repetitive answers with similar content. 7) Sometimes, hallucinations occur (details in material \& device prediction with LLM section).

In comparison, fine-tuning exhibits significant advantages: 1) Top experts in the field design the framework and aligns more closely with the domain-specific experimental thinking;
2) it saves the cost, time, and effort of annotation; 3) professionals in the relevant field can directly use the results of the same framework without any additional learning costs.

\textbf{Material \& Device Prediction with LLM}

Upon further investigation of the results, we identify a phenomenon known as "hallucination" within the large language model (LLM) outputs in Set \textit{D}. In this context, "hallucination" refers to instances where no stability test is mentioned in the input, but the fine-tuned SII model is employed. We devise two tasks to quantify the model’s performance: regression and classification tasks for predicting performance data. However, as only 11\% of the training device data has undergone stability tests, the sample size is insufficient to generate adequate training and test sets. Consequently, we opt for electrical performance data, as all data points possess associated values, including open-circuit voltage, V\textsubscript{oc}, short-circuit current, J\textsubscript{sc}, fill factor, FF, and power conversion efficiency, PCE. Notably, the model only predicts data points under JV\_light\_spectra under AM1.5 and JV\_light\_intensity equal to 1000 W/m\textsuperscript{2}.

\textbf{Classification:} Our first task entails a classification challenge aimed at evaluating the ability of the fine-tuned GPT to accurately identify the PCE level of perovskite solar cells under the AM1.5 spectrum and 1000 W/m\textsuperscript{2} light intensity, considering specific parameters such as stack and method information. We categorize the PCE levels into four groups: low  (0\%-8\%), normal (8\%-18\%), and high (>18\%). We expect the model to generate text completions corresponding to these four categories. Appendix C offers a comprehensive example schema outlining the structure and synthesis of a perovskite solar cell. Figure 5 illustrates the performance of the fine-tuned GPT model in the MDP task.

\begin{figure}[h]
  \centering
  \includegraphics[width=0.5\linewidth]{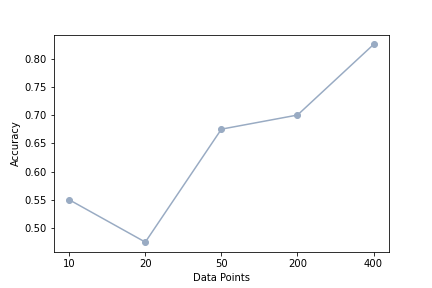}
  \caption{Peformance on the Classification Tasks}
\end{figure}

\textbf{Regression:} A more demanding task than classification involves developing a regression model capable of predicting continuous property values, including V\textsubscript{oc}, J\textsubscript{sc}, FF, and PCE, for perovskite solar cells employing specific synthesis methods. Although large language models cannot predict real numbers in highly precise regression tasks, they can still produce predictions of acceptable accuracy by employing rounded values during training. A precision of two decimal points is deemed sufficient for electrical performance data. Appendix C presents a detailed example schema. Table 4 and Figure 6 demonstrate the performance of the fine-tuned GPT model in MDP for regression tasks.

\textbf{Potential of LLM in MDP tasks}

Our study demonstrates that an LLM, even without prior training in materials science, can predict device performance data that may not be explicitly stated in the literature. Although the generated hallucinated information is not completely accurate, it remains valuable for researchers using the amassed scientific knowledge. In contrast to the recent advancements in perovskite solar cell prediction by Liu et al. \cite{Liu2022HowCells}, who manually collected 814 data points from 2,735 publications and built machine-learning models for J-V performance prediction using only 13 features, LLMs are capable of automatically generating higher-dimensional datasets. This ability allows LLMs to guide subsequent device design at the material level, accounting for parameters such as annealing time, annealing temperature, material thickness, and area, with greater flexibility in feature selection. Feature values are not limited to numbers and are readily obtainable by scientists. Jablonka et al.\cite{Jablonka2023Is} also demonstrated that GPT-3 performs comparably or outperforms traditional techniques when confronted with limited data, particularly for organic compounds with unique line encodings like SMILES \cite{Weininger1988SMILESRules} or SELFIES \cite{Krenn2022SELFIESRepresentations}. Similarly, we devised a schema for predicting the organic photovoltaic devices (OPV) PCE (density functional theory, DFT, calculated) based on the Harvard Photovoltaic (HOPV15) \cite{Lopez2016TheDataset} Dataset. Compared to the BRANNLP (Bayesian Regularized Artificial Neural Network with Laplacian prior) method employed by Meftahi et al.\cite{Meftahi2020MachineDevices}, fine-tuned GPT achieves comparable performance with a simple schema design (details could be found in Appendix D).

\begin{table}[h!]
\centering
\caption{Mean Square Errors(MAE) of Regression tasks for prediction of the electrical performance of perovskite solar cell}
    \begin{tabular}{lllllll} 
    \toprule
    \multicolumn{6}{c}{\bfseries MAE } \\
    \cmidrule(lr){2-6}
    Sample  & 10 & 20 & 50 & 100 & 400 \\ 
    \midrule
    \multirow{1}{*}{J\textsubscript{sc}} & 7.62  & 7.84 & 6.38 & 5.15 & 3.59 \\ 
    \midrule
    \multirow{1}{*}{V\textsubscript{oc}} & 0.17  & 0.18 & 0.12 & 0.09 & 0.104  \\ 
    \midrule
    \multirow{1}{*}{FF} & 0.11 & 0.12 & - & - & 0.105  \\ 
    \midrule
    \multirow{1}{*}{PCE} & 8.47\%  & 5.21\% & 3.48\% & 4.05\% & 2.61\% \\
    \bottomrule
    \end{tabular}
\end{table}

\begin{figure}[h]
  \centering
  \includegraphics[width=0.8\textwidth]{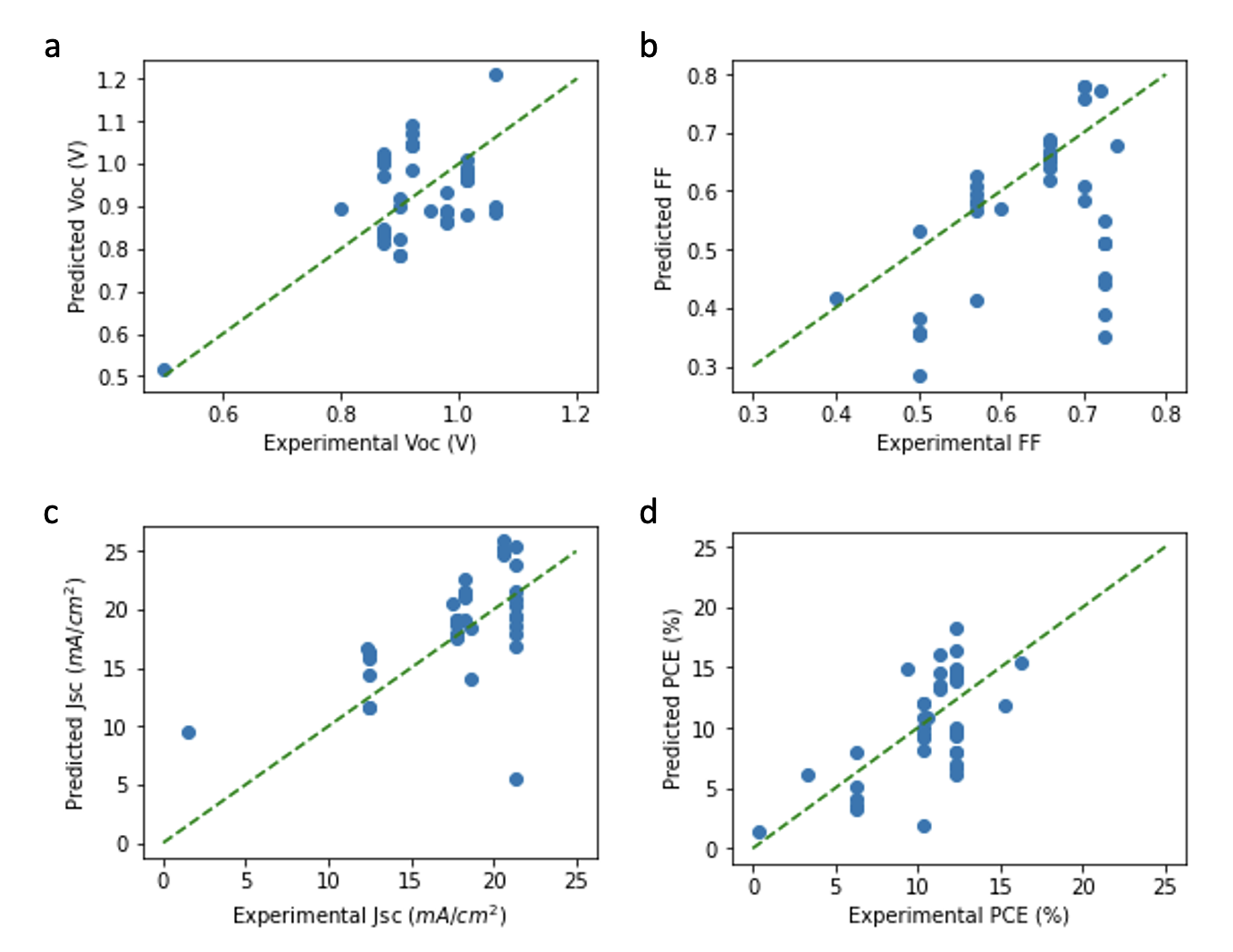}
  \caption{Comparison of experimental values and Fine-tuned GPT-3 prediction values.}
\end{figure}

LLM, such as the GPT-3 used in the study, have demonstrated an aptitude for identifying structural and property-related similarities between novel materials and those previously investigated, akin to the expertise of seasoned materials scientists. This capability to discern similarities enables the exploration of variations in these new materials, consequently opening up opportunities for innovative applications. Moreover, LLMs exhibit the potential to design cutting-edge devices by harnessing detailed material information. Although GPT-3 was not originally tailored for scientific fields, its performance in this realm suggests a promising future for LLMs in scientific tasks. By augmenting LLMs with further training in relevant scientific literature, they may ultimately be empowered to guide experimental design and substantially broaden their scope of applications in materials science and beyond.

\textbf{Limitation}

In this study, we observed that failures predominantly transpire when a sample surpasses GPT-3's prompt-completion token limit, which was set at 2048 during the investigation. This limitation implies that paragraphs characterized by considerable length or high information density are fundamentally incompatible with the current approach. A significant proportion of unparseable completions can be attributed to instances where the passage and partial completion extend beyond the imposed token limit, consequently leading to premature truncation and hindering the generation of a comprehensive output.

\section{Conclusion}

In this study, we introduce a new NLP task, structured information inference (SII), which aims to obtain hierarchical, domain-specific material and device information from unstructured scientific texts. After analyzing the traditional annotation mechanisms and characteristics of LLM, we proposed to solve this task by finetuning one of LLMs, GPT-3. Remarkably, this approach does not necessitate manual annotation, instead relying on review papers or FAIR datasets for training purposes. By employing this method, GPT-3 can effectively predict material properties and device performance, as well as generate innovative materials or devices tailored to meet specialized requirements. Demonstrating exceptional flexibility, the approach readily adapts to various challenges within scientific domains and exhibits outstanding performance in both SII and MDP tasks, particularly for perovskite and organic photovoltaic devices. Existing LLMs, such as GPT-3, can leverage this method to extract structured relational datasets, thereby guiding material development. The strategy could also be applied to LLMs with encoder-decoder architecture like GPT-3, including LLaMA and Palm-E. This end-to-end approach ultimately seeks to empower scientists with the ability to swiftly generate material knowledge and design novel materials or chemicals for research purposes. To showcase the method, an online demonstration can be accessed at http://www.masterai.com.au.
\section{Acknowledgements}

T.X expresses gratitude for the PhD scholarship provided by UNSW Future Manufacturing Institute. study was also undertaken with the assistance of resources and services from the National Computational Infrastructure (NCI), which is supported by the Australian Government, and through the UNSW-NCI partner trial scheme. We thank C. Siu from the UNSW Library for assisting our team in obtaining Text and Data Mining agreements with publishers. Additionally, we acknowledge H. Wen from Carnegie Mellon University for valuable discussions on developing the SII task, M. He and Y. Wang from University of New South Wales for SII schema design and L. Duan from Australian National University for MDP schema design.

\bibliographystyle{unsrt}

\clearpage
\appendix
\section{Exact match rate of elements}
\label{sec:appendix1}
\begin{table}[h!]
\centering
    \begin{tabular}{ccc}
    \toprule
    Element name & Exact match rate & Support\\
    \midrule
    Substrate\_stack\_sequence & 47.58 & 2270\\
    ETL\_stack\_sequence & 3.95 & 2277\\
    ETL\_additives\_compounds & 68.73 & 266\\
    ETL\_deposition\_procedure & 50.99 & 2270\\
    Perovskite\_composition\_long\_form & 22.40 & 2277\\
    Perovskite\_composition\_short\_form & 21.87 & 2277\\
    Perovskite\_additives\_compounds & 82.84 & 669\\
    Perovskite\_deposition\_solvents & 73.17 & 2218\\
    Perovskite\_deposition\_procedure & 51.33 & 2270\\
    Perovskite\_deposition\_thermal\_annealing\_temperature & 73.41 & 2069\\
    Perovskite\_deposition\_thermal\_annealing\_time & 60.65 & 2015\\
    HTL\_stack\_sequence & 39.86 & 2275\\
    HTL\_additives\_compounds & 43.26 & 1185\\
    HTL\_deposition\_procedure & 50.62 & 2270\\
    Backcontact\_stack\_sequence & 68.33 & 2270\\
    Backcontact\_additives\_compounds & 4.17 & 72\\
    Backcontact\_deposition\_procedure & 47.27 & 2270\\
    Stability\_measured & 9.49 & 2277\\
    Stability\_average\_over\_n\_number\_of\_cells & 0 & 0\\
    Stability\_temperature\_range & 0 & 0\\
    Stability\_atmosphere & 14.14 & 2270\\
    Stability\_time\_total\_exposure & 43.60 & 523\\
    Stability\_PCE\_initial\_value & 13.64 & 220\\
    Stability\_PCE\_end\_of\_experiment & 64.30 & 521\\
    Cell\_area\_measured & 72.61 & 2227\\
    Cell\_number\_of\_cells\_per\_substrate & 0 & 0\\
    Cell\_architecture & 0 & 0\\
    Cell\_flexible & 0 & 0\\
    Cell\_semitransparent & 0 & 0\\
    Cell\_semitransparent\_wavelength\_range & 0 & 0\\
    Module & 0 & 0\\  

  \bottomrule
\end{tabular}
\end{table}

\section{Division of Relation Sets}
\label{sec:appendix}
Set A: \\
'Substrate\_stack\_sequence', \\
'ETL\_stack\_sequence',\\
'ETL\_additives\_compounds',\\
'Perovskite\_composition\_long\_form', \\
'Perovskite\_composition\_short\_form', \\
'Perovskite\_additives\_compounds',\\
'HTL\_stack\_sequence',\\
'HTL\_additives\_compounds', \\
'Backcontact\_stack\_sequence',\\
'Backcontact\_additives\_compounds'\\
\\
Set B:\\
'Cell\_area\_measured',\\ 
'Cell\_number\_of\_cells\_per\_substrate', \\
'Cell\_architecture', \\
'Cell\_flexible', \\
'Cell\_semitransparent', \\
'Cell\_semitransparent\_wavelength\_range'\\
\\         
Set C:\\
'ETL\_deposition\_procedure', \\
'HTL\_deposition\_procedure',\\
'Backcontact\_deposition\_procedure',\\
'Perovskite\_deposition\_procedure',\\
'Perovskite\_deposition\_solvents', \\
'Perovskite\_deposition\_thermal\_annealing\_temperature',\\
'Perovskite\_deposition\_thermal\_annealing\_time'\\
\\
Set D:\\
'Stability\_measured', \\
'Stability\_average\_over\_n\_number\_of\_cells',\\ 
'Stability\_temperature\_range', \\
'Stability\_atmosphere',\\ 
'Stability\_time\_total\_exposure', \\
'Stability\_PCE\_initial\_value', \\
'Stability\_PCE\_end\_of\_experiment'\\

\section{MDP Example Schema}
\textbf{Classification}

Prompts : 

What's the PCE of the perovskite solar cell with the parameters below:\\
Substrate\_stack\_sequence', \\
'ETL\_stack\_sequence',\\
'ETL\_additives\_compounds',\\
'Perovskite\_composition\_long\_form', \\
'Perovskite\_composition\_short\_form', \\
'Perovskite\_additives\_compounds',\\
'HTL\_stack\_sequence',\\
'HTL\_additives\_compounds', \\
'Backcontact\_stack\_sequence',\\
'Backcontact\_additives\_compounds'\\
'ETL\_deposition\_procedure', \\
'HTL\_deposition\_procedure',\\
'Backcontact\_deposition\_procedure',\\
'Perovskite\_deposition\_procedure',\\
'Perovskite\_deposition\_solvents', \\
'Perovskite\_deposition\_thermal\_annealing\_temperature',\\
'Perovskite\_deposition\_thermal\_annealing\_time'?\\

\textbf{Regression}

Prompts : 

What's the PCE value of the perovskite solar cell with the parameters below:\\
Substrate\_stack\_sequence', \\
'ETL\_stack\_sequence',\\
'ETL\_additives\_compounds',\\
'Perovskite\_composition\_long\_form', \\
'Perovskite\_composition\_short\_form', \\
'Perovskite\_additives\_compounds',\\
'HTL\_stack\_sequence',\\
'HTL\_additives\_compounds', \\
'Backcontact\_stack\_sequence',\\
'Backcontact\_additives\_compounds'\\
'ETL\_deposition\_procedure', \\
'HTL\_deposition\_procedure',\\
'Backcontact\_deposition\_procedure',\\
'Perovskite\_deposition\_procedure',\\
'Perovskite\_deposition\_solvents', \\
'Perovskite\_deposition\_thermal\_annealing\_temperature',\\
'Perovskite\_deposition\_thermal\_annealing\_time'?\\

\section{OPV Schema and Performance}
\textbf{Schema Design}

Example Prompts: what is the PCE of donors: COC(=O)c1cc2csc(-CC1(C)c2ccsc2-c2sc(-c3c4nsnc4cc(F)c3F)cc21, and Acceptors: PC71BM?

Completion: 2.00

\textbf{Model Development}

Both the BRANNLP and GPT-3 model was trained on 276 donor-acceptors pairs and tested on 68 donor-acceptors. The difference is in how to encode the donor and acceptors,

BRANNLP: donor encoded by signature descriptors, and acceptor encoded by '1-hot' indicator variable. (details in OPV datasets).

GPT-3: donor encoded by SMILE system and acceptor encoded by their IUPAC name.

\textbf{Model Performance}
\begin{figure}[h]
  \centering
  \includegraphics[width=0.5\linewidth]{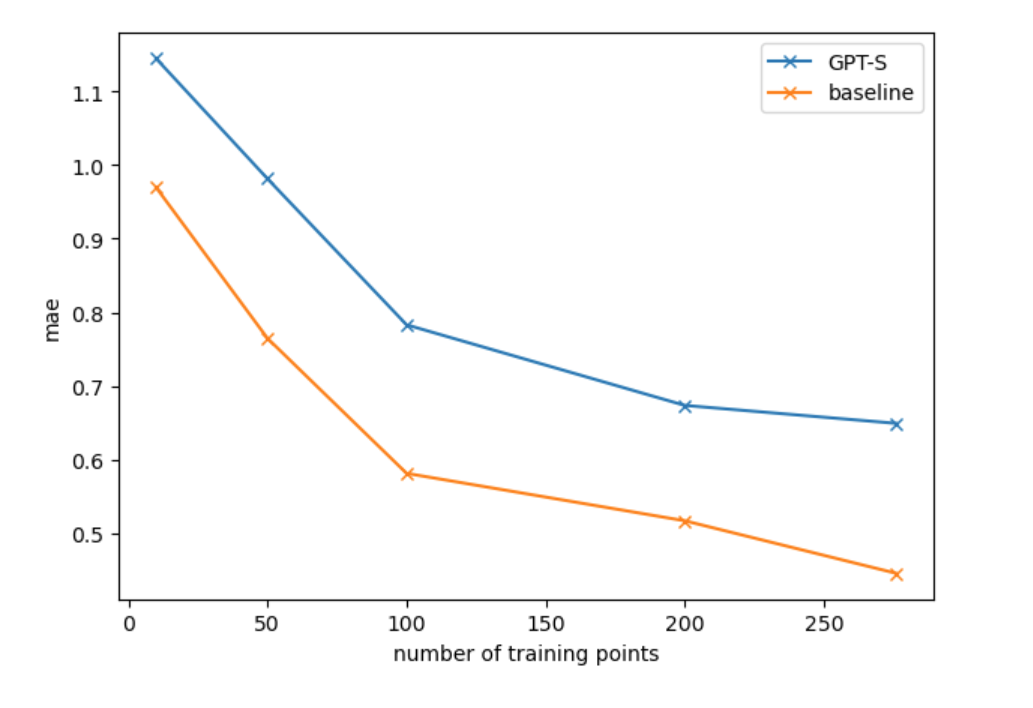}
  \caption{Peformance of GPT-SMILE(GPT-S) and BRANNLP(baseline) models in prediction OPV PCE(\%), mae is the mean absolute errors. }
\end{figure}

\end{document}